%% file: main.tex
\documentclass{bmvc2k}
\input{math_commands.tex}

\usepackage{times}
\usepackage{epsfig}
\usepackage{graphicx}
\usepackage{amsmath}
\usepackage{amssymb}
\usepackage{mathtools}
\usepackage{pifont}
\newcommand{\cmark}{\text{\ding{51}}}
\newcommand{\xmark}{\text{\ding{55}}}
\usepackage{booktabs}
\usepackage{amsthm}
\usepackage{amsfonts}
\usepackage{color}
\usepackage{float}
\usepackage{multirow}
\usepackage{algorithmicx}
\usepackage{algorithm}
\usepackage[normalem]{ulem}
\usepackage{algpseudocode}

\newcommand{\vect}[1]{{{#1}}}

\DeclarePairedDelimiterX{\infdivx}[2]{(}{)}{%
  #1\;\delimsize|\delimsize|\;#2%
}
\newcommand{\kld}[2]{\ensuremath{D_{\mathrm{KL}}\infdivx{#1}{#2}}\xspace}

\newcommand{\vP}[0]{\vect{p}}

\newcommand{\thetaa}[0]{\vect{\theta_1}}
\newcommand{\thetab}[0]{\vect{\theta_2}}
\newcommand{\vthetai}[1]{\vect{\theta_{#1}}}
\newcommand{\x}[0]{\vect{x}}
\newcommand{\xadv}[1]{\vect{x_\mathrm{adv}^{#1}}}
\newcommand{\xadvm}[2]{\vect{x_\mathrm{adv_{#1}}^{#2}}}
\newcommand{\loss}[1]{\mathcal{L}_{\operatorname{#1}}}
\newcommand{\attack}[0]{\texttt{attack}}

\algnewcommand{\LineComment}[1]{\State \(\triangleright\) #1}

\title{Mutual Adversarial Training:\\ Learning together is better than going alone.}

\addauthor{Jiang Liu}{jiangliu@jhu.edu}{1}
\addauthor{Chun Pong Lau}{clau13@jhu.edu}{1}
\addauthor{Hossein Souri}{hsouri1@jhu.edu}{1}
\addauthor{Soheil Feizi}{sfeizi@cs.umd.edu}{2}
\addauthor{Rama Chellappa}{rchella4@jhu.edu}{1}
\addinstitution{
 Johns Hopkins University,\\
 Baltimore, Maryland, USA
}
\addinstitution{
 University of Maryland, College Park\\
 College Park, Maryland, USA
}

\runninghead{Jiang Liu}{Mutual Adversarial Training}

\def\eg{\emph{e.g}\bmvaOneDot}

\begin{document}

\maketitle

\begin{abstract}
Recent studies have shown that robustness to adversarial attacks can be transferred across networks. In other words, we can make a weak model more robust with the help of a strong teacher model. We ask if instead of learning from a static teacher, can models “learn together” and “teach each other” to achieve better robustness? In this paper, we study how interactions among models affect robustness via knowledge distillation. We propose mutual adversarial training (MAT), in which multiple models are trained together and share the knowledge of adversarial examples to achieve improved robustness. MAT allows robust models to explore a larger space of adversarial samples, and find more robust feature spaces and decision boundaries. Through extensive experiments on CIFAR-10 and CIFAR-100, we demonstrate that MAT can effectively improve model robustness and outperform state-of-the-art methods under white-box attacks, bringing $\sim$8\% accuracy gain to vanilla adversarial training (AT) under PGD-100 attacks. In addition, we show that MAT can also mitigate the robustness trade-off among different perturbation types, bringing as much as 13.1\% accuracy gain to AT baselines against the union of $l_\infty$, $l_2$ and $l_1$ attacks. These results show the superiority of the proposed method and demonstrate that collaborative learning is an effective strategy for designing robust models.
\end{abstract}

\section{Introduction}
\label{sec:intro}
In recent years, we have witnessed the great success of deep neural networks (DNNs) in various fields of artificial intelligence including computer vision~\cite{krizhevsky2012imagenet}, speech recognition~\cite{6296526}, and robot control~\cite{levine2018learning}. Despite their superior performance, DNNs are shown to be vulnerable to adversarial attacks that add imperceptible manipulations to inputs~\cite{madry2017towards, goodfellow2014explaining,laidlaw2019functional, laidlaw2020perceptual, kang2019testing, dong2018boosting, carlini2017towards, chen2017ead, 7467366}. This poses a huge challenge in security critical applications such as autonomous driving and medicine.

To enhance model robustness against adversarial attacks, many defense methods have been proposed including empirical~\cite{madry2017towards,zhang2019theoretically, papernot2016distillation,lin2020dual, singla2021skew, Kannan2018AdversarialLP, wang2019improving} and certifiable defenses~\cite{raghunathan2018certified,cohen2019certified, levine2020randomized, levine2019certifiably, sinha2017certifying, levine2021improved}. Adversarial training (AT)~\cite{madry2017towards} is considered to be one of the most effective algorithms for adversarial defenses. There have been many works for improving adversarial training by using different loss functions, such as ALP~\cite{Kannan2018AdversarialLP}, TRADES~\cite{zhang2019theoretically}, and MART~\cite{wang2019improving}. However, they only train one model without considering the synergy of a network cohort. In addition, most defense methods focus on a single perturbation type, which can be vulnerable to unseen perturbation types ~\cite{tramer2019adversarial, maini2020adversarial}. For example, models adversarially trained on $l_\infty$-bounded adversarial examples can be vulnerable to $l_1$ and $l_2$-bounded attacks. 

Knowledge distillation (KD)~\cite{hinton2015distilling} is a well-known method for transferring knowledge learned by one model to another. There are many forms of KD~\cite{wang2020knowledge}  including \textit{offline} KD, where the teacher models are pretrained and the students learns from static teachers\cite{hinton2015distilling}, and \textit{online} KD, where a group of student models learn from peers' predictions~\cite{zhang2018deep, guo2020online, song2018collaborative}. Several techniques based on KD have been proposed for adversarial defenses~\cite{papernot2016distillation, goldblum2019adversarially, arani2020adversarial, chen2021robust}. ~\cite{goldblum2019adversarially} demonstrates that robustness can be transferred among models through KD and~\cite{chen2021robust} shows that KD can help mitigate robust overfitting. Table~\ref{tab:kd} summarizes the differences among KD-based defenses. We argue that current KD-based defenses are sub-optimal in terms of improving adversarial robustness. Defensive distillation (DD) trains a natural model with another natural model as the teacher, which cannot provide strong robustness as both the teacher and student are not adversarially trained, and it is broken by~\cite{carlini2016defensive}. Adversarially robust distillation (ARD)~\cite{goldblum2019adversarially} trains a robust model with another robust model as the teacher to distill the robustness of a large network onto a smaller student, which relies on the existence of strong teacher models, and the improvement of the student model is limited as the teacher is fixed. Adversarial concurrent training (ACT)~\cite{arani2020adversarial} trains a natural and a robust model jointly in an online KD manner to align the feature space of both. However, since natural models and robust models learn fundamentally different features~\cite{tsipras2018robustness}, aligning the feature space of a robust model with a natural model may result in degraded robustness.

\begin{table}
\caption{Comparisons of KD-based defenses. "Robust" means the model is trained with adversarial examples and "Natural" means the model is trained with natural examples.}
\label{tab:kd}
\centering
\small
\scalebox{0.9}{
\begin{tabular}{c|c|c|c|c}
\hline
Method                 & Teacher Model & Student Model & Form of KD & Multi-Perturbations  \\\hline
DD~\cite{papernot2016distillation} & Natural    & Natural    & Offline    & \xmark                       \\
ARD~\cite{goldblum2019adversarially}                    & Robust      & Robust      & Offline    & \xmark                       \\
ACT~\cite{arani2020adversarial}          & Natural    & Robust      & Online     & \xmark                       \\
MAT \textbf{(Ours)}                    & Robust      & Robust      & Online     & \cmark   \\\hline                  
\end{tabular}}
\end{table}

\begin{figure}
  \centering
  \includegraphics[width=0.9\textwidth]{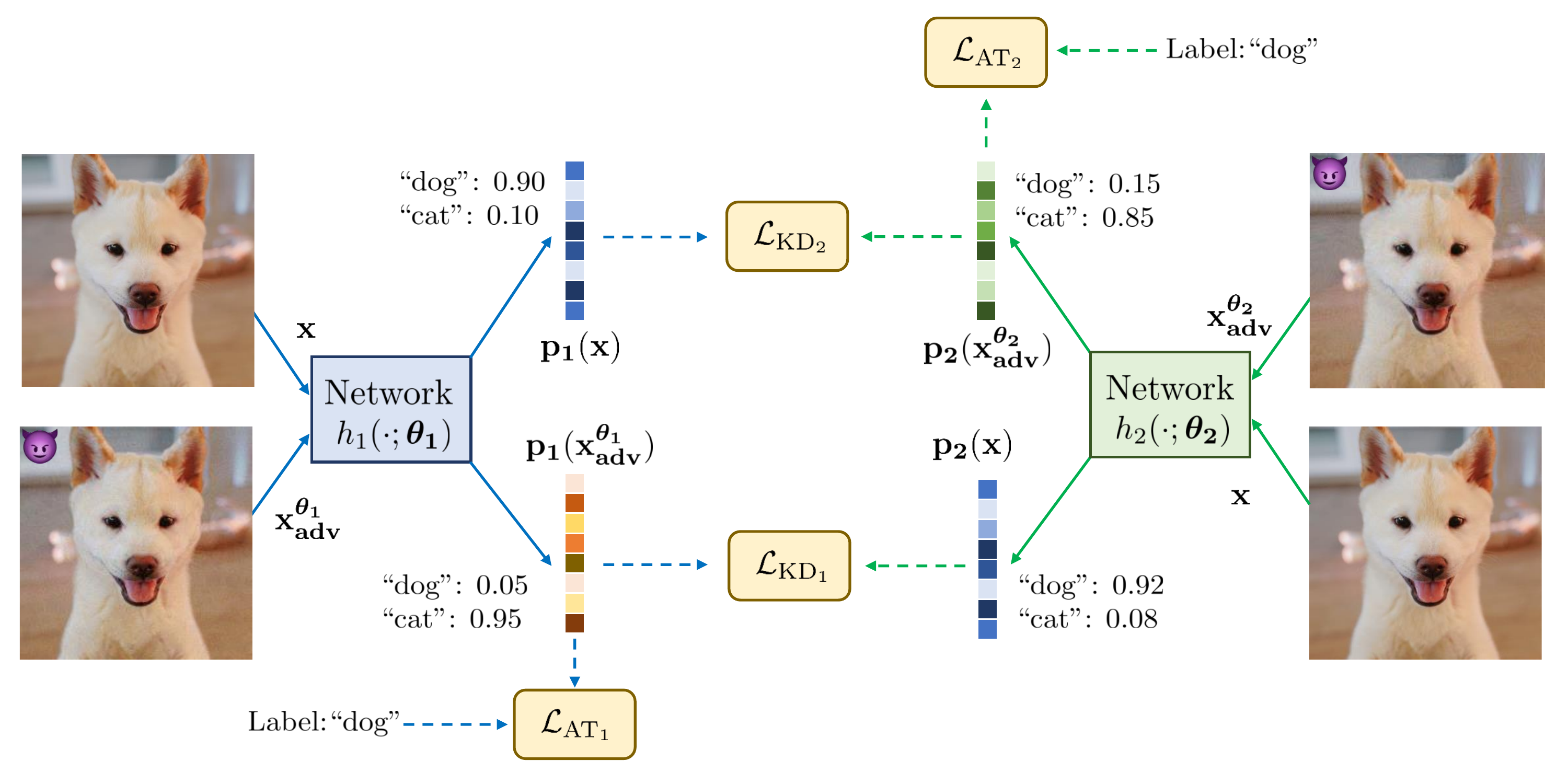}
  \caption{Mutual adversarial training (MAT) architecture. $\x$ is a clean image, $\xadv{\thetaa}$ is the adversarial image of network $h_1$, and $\xadv{\thetab}$ is the adversarial image of network $h_2$. } 
  \label{fig:mat}
\end{figure}

In this paper, we propose Mutual Adversarial Training (MAT) that allows models to share their knowledge of adversarial robustness and teach each other to be more robust. Unlike previous KD-based defenses, we train a group of robust models simultaneously, and each network not only learns from ground-truth labels as in standard AT, but also the soft labels from peer networks that encode peers’ knowledge for defending adversarial attacks to achieve stronger robustness. The architecture of MAT is shown in Fig.~\ref{fig:mat}. 

MAT improves model robustness through many aspects: 1) MAT inherits the benefits of KD, such as improving generalization~\cite{furlanello2018born} and reducing robust overfitting~\cite{chen2021robust}. 2) \textit{MAT creates a positive feedback loop of increasing model robustness}. Each network serves as a robust teacher to provide semantic-aware and discriminative soft labels to its peers. By learning from strong peers, a network  becomes  more  robust, which in turn improves the robustness of its peers. 3) \textit{MAT allows robust models to explore a larger space of adversarial samples, and find more robust feature spaces and decision boundaries jointly}. The adversarial examples of each model forms a subspace of adversarial inputs~\cite{tramer2017space}, and the predictions of each model encode information about its decision boundary and adversarial subspace. In MAT, each model not only learns from its own adversarial examples, but also receives information about peers' adversarial examples through the soft labels. In this way, each model needs to consider a larger space of adversarial samples, and find a feature space and decision boundary that not only work well on its own adversarial examples but also on peers' adversarial examples, which encourages solutions that are more robust and generalizable. 

To summarize, in this paper we propose a novel KD-based adversarial training algorithm named MAT. MAT is a general framework for boosting adversarial robustness of any network without the need for strong teacher models.  We further extend MAT for defending against multiple perturbation types (MAT-MP) by exploiting the transferability of adversarial robustness, and propose several training strategies for training MAT-MP. Our extensive experiments show that MAT brings significant robustness improvements to AT baselines and outperforms state-of-the-art methods for both single and multiple perturbations.

\section{Mutual Adversarial Training}

\subsection{Our Framework} 

In this paper, we consider the $K$-class $(K\geq 2)$ image classification problem. We have image and label tuples $(\vect{x}, y)$ drawn from an underlying data distribution $\mathcal{D}$, where $\vect{x} \in \sR^d$ is a natural image, $y=\{1, \cdots, K\}$ is its corresponding class label. We formulate the proposed mutual adversarial training algorithm with a cohort of two networks $h_1(\cdot;\thetaa)$ and $h_2(\cdot;\thetab)$. 

To train a robust $h_1$, we minimize the classification loss on the adversarial examples as in adversarial training. 
The AT loss is then defined as:
\begin{equation}
    \label{eq:at1}
    \loss{AT_1}=\loss{C}(\vect{p_1}(\xadv{\thetaa}), y),
\end{equation}
where $\vect{p_1}$ is the output probability of $h_1$, $\loss{C}$ is a classification loss, and $\xadv{\thetaa}$ is an adversarial example crafted for $h_1$, $\xadv{\thetaa}=\x+\argmax_{\vect{\delta}\in S}\loss{C}(\vect{p_1}(\x+\vect{\delta}), y)$. For $\loss{C}$, we use the boosted cross entropy loss as in MART~\cite{wang2019improving}, which uses a margin term to improve the decision margin of the classifier:
$\loss{C}(\vP(\x), y)=-\log(\vP(\x)_y)-\log(1-\max_{k\neq y}\vP(\x)_k)$.

To boost the robustness of $h_1$, we use the peer network $h_2$ to provide its knowledge of robustness to $h_1$. Specifically, we use the KD loss to guide the prediction of $h_1$:
\begin{equation}
    \loss{KD_1}=\kld{\vect{p_2}(\x)}{\vect{p_1}(\xadv{\thetaa})},
\end{equation}
where $\kld{\cdot}{\cdot}$ is the Kullback–Leibler (KL) divergence and $\vect{p_2}$ is the output probability of $h_2$. Note that we use the clean image $\x$ as the input to $h_2$ to generate the soft label $\vect{p_2}$ because adversarial examples can mislead $h_2$ to generate wrong predictions. The KD loss aligns the feature spaces and decision boundaries of MAT models and allows them to find robust features and decision boundaries jointly. Similarly, we define $\loss{AT_2}$ and $\loss{KD_2}$ for $h_2$:
\begin{equation}
    \loss{AT_2}=\loss{C}(\vect{p_2}(\xadv{\thetab}), y),\ 
    \loss{KD_2}=\kld{\vect{p_1}(\x)}{\vect{p_2}(\xadv{\thetab})},
\end{equation}
where $\xadv{\thetab}$ is an adversarial example crafted for $h_2$, $\xadv{\thetab}=\x+\argmax_{\vect{\delta}\in S}\loss{C}(\vect{p_2}(\x+\vect{\delta}), y)$.
The overall loss function for MAT is:
\begin{equation}
    \label{eq:mat}
    \loss{MAT} = (1-\alpha)(\loss{AT_1}+\loss{AT_2})+\alpha(\loss{KD_1} + \loss{KD_2}),
\end{equation}
where $\alpha$ is a hyper-parameter that controls the trade-off between learning from ground-truth labels and the peer network. The MAT algorithm for the case of two classifiers is summarized in Alg.~\ref{alg:mat}.

The extension of MAT to larger student cohorts is straightforward. Consider $N\ (N\geq2)$ networks $h_n(\cdot;\vect{\theta_n}) (n=1,\cdots, N)$, the loss function of MAT becomes:
\begin{equation}
    \label{eq:mat-loss}
    \loss{MAT}=(1-\alpha)\loss{AT}+\frac{\alpha}{M-1}\loss{KD},
\end{equation}
where
    $\loss{AT}=\sum_{n=1}^N\loss{C}(\vect{p_n}(\xadv{\vect{\theta_n}}), y)$, and
    $\loss{KD}=\sum_{n=1}^N\sum_{m\neq n}^N\kld{\vect{p_m}(\x)}{\vect{p_n}(\xadv{\vthetai{n}})}$.
\begin{algorithm}[t]
\begin{algorithmic}[1]
\Require Training samples $\gX \times \gY$, classifier $h_1(\cdot;\thetaa)$ and $h_2(\cdot;\thetab)$, attack model $\attack$, learning rate $\tau$, weight $\alpha$.
\State Randomly initialize $\thetaa$, $\thetab$
\For{epoch = 1, \dots, N}
    \For{minibatch $(\vect{x}, y) \subseteq \gX \times \gY$}
        \State $\xadv{\thetaa}=\attack(\x, y, h_1(\cdot;\thetaa))$ 
        \LineComment{Generate adversarial examples of $h_1$}
        \State $\xadv{\thetab}=\attack(\x, y, h_2(\cdot;\thetab))$ 
        \LineComment{Generate adversarial examples of $h_2$}
        \State $\loss{MAT} = (1-\alpha)(\loss{AT_1}+\loss{AT_2})+\alpha(\loss{KD_1} + \loss{KD_2})$ \Comment{Compute loss}
        \State $\thetaa \gets \thetaa -\tau\nabla_{\thetaa}\loss{MAT}$
        \State $\thetab \gets \thetab -\tau\nabla_{\thetab}\loss{MAT}$
        \LineComment{Update $\thetaa$ and $\thetab$ with gradient descent}
    \EndFor
\EndFor
\end{algorithmic}
\caption{Mutual Adversarial Training (MAT)}
\label{alg:mat}  
\end{algorithm}

\subsection{Defense against Multiple Perturbations}
In this section, we demonstrate how MAT can improve robustness to multiple perturbations. Previous methods~\cite{tramer2019adversarial, maini2020adversarial} attempt to improve model robustness against multiple perturbations by augmenting the training data. We take a different approach by transferring the robustness of specialist models to a single model, which complements previous methods and achieves better performance.   

Given $M$ $(M\geq 2)$ different perturbation types, since it is difficult for a single model to be robust to all perturbations, we train an ensemble of $M+1$ networks including one generalist network $h_{0}(\cdot;\vthetai{0})$ and $M$ specialist networks $h_1(\cdot;\thetaa), h_2(\cdot;\thetab), \cdots, h_{M}(\cdot;\vthetai{M})$. Each specialist network is responsible for learning to defend against a specific perturbation and learns from peer networks to boost its robustness. The loss for training specialist networks $h_m\ (1\leq m\leq M)$ is defined as:
\begin{equation}
    \label{eq:Lm}
          \loss{m}=(1-\alpha)\loss{C}(\vect{p_m}(\xadvm{m}{\vthetai{m}}), y) + \frac{\alpha}{M}\sum_{n\neq m}^{M}\kld{\vect{p_n}(\x)}{\vect{p_m}(\xadvm{m}{\vthetai{m}})},   
\end{equation}
where $\xadvm{m}{\vthetai{m}}$ is an adversarial example of $h_m$ generated by the $m$-th attack model. 

The generalist network $h_0$ learns to defend against all perturbations with the help of specialist networks. Specifically, when classifying adversarial examples generated by the $m$-th attack model, the generalist compares its prediction with ground-truth labels, as well as the predictions of the $m$-th specialist model $h_m$ that specializes in this attack. In this way, the generalist is able to consider the decision boundaries of different perturbations and find an optimal one that is robust to the union of multiple perturbations. 

We propose three strategies for training the generalist $h_0$: 

1) MAT-AVG: we train $h_0$ with the average loss on all perturbations at each iteration.  The loss for training the generalist network $h_0$ in MAT-AVG is defined as:
\begin{equation}
    \label{eq:L-avg}
          \loss{0}=\frac{1}{M}\sum_{m=1}^M\left[ (1-\alpha)\loss{C}(\vect{p_0}(\xadvm{m}{\vthetai{0}}), y) +
          \alpha \kld{\vect{p_m}(\x)}{\vect{p_0}(\xadvm{m}{\vthetai{0}})}\right],  
\end{equation}
where $\xadvm{m}{\vthetai{0}}$ is an adversarial example of $h_0$ generated by the $m$-th attack model.  

2) MAT-MAX: we train $h_0$ with the worse perturbation only. The loss for training the generalist network $h_0$ in MAT-MAX is defined as:
\begin{equation}
    \label{eq:L-max}
          \loss{0}= (1-\alpha)\loss{C}(\vect{p_0}(\xadvm{k}{\vthetai{0}}), y) +\alpha\kld{\vect{p_k}(\x)}{\vect{p_0}(\xadvm{k}{\vthetai{0}})},   
\end{equation}
where $\operatorname{k}=\argmax_{\operatorname{m}}\loss{C}(\vect{p}(\xadvm{m}{\vthetai{0}}), y)$.

3) MAT-MSD: we train $h_0$ using MSD~\cite{maini2020adversarial} adversarial examples. Since MSD incorporates multiple perturbation models into a single attack, $h_0$ learns from all the specialists at each iteration. The loss for training the generalist network $h_0$ in MAT-MSD is defined as:
\begin{equation}
    \label{eq:L-msd}
          \loss{0}=\frac{1}{M}\sum_{m=1}^M\left[ (1-\alpha)\loss{C}(\vect{p_0}(\x_{\text{MSD}}^{\vthetai{0}}), y) +
          \alpha \kld{\vect{p_m}(\x)}{\vect{p_0}(\x_{\text{MSD}}^{\vthetai{0}})}\right],  
\end{equation}
where $\x_{\text{MSD}}^{\vthetai{0}}$ is the MSD adversarial example generated for $h_0$.

The total loss of Mutual Adversarial Training for Multiple Perturbations (MAT-MP) is the sum of all networks:
\begin{equation}
    \loss{MAT-MP}=\sum_{m=0}^{M}\mathcal{L}_{m}.
\end{equation}

\section{Experiments and Results}
\subsection{Implementation Details}
We evaluate the robustness of the proposed methods on CIFAR-10 and CIFAR-100~\cite{krizhevsky2009learning}. All images are normalized into [0, 1], and data augmentations are used during training including random horizontal flipping and 32$\times$32 random cropping with 4-pixel padding. We use ResNet-18~\cite{he2016deep} and WideResNet-34-10 (WRN-34-10)~\cite{BMVC2016_87} networks and train them with SGD optimizers with a batch size of 128 for 120 epochs. For ResNet-18, we set the initial learning rate to 0.01, momentum to 0.9 and weight decay to $3.5\times 10^{-3}$. For WRN-34-10, we set the initial learning rate to 0.1, momentum to 0.9 and weight decay to $7\times 10^{-4}$. The learning rate drops by 0.1 at the 75-th, 90-th and 100-th epochs.   
\subsection{Defense against Single Perturbation}
\subsubsection{Settings}
In this section, we evaluate the effectiveness of MAT against a single perturbation. We consider the popular $l_\infty$ attacks with $\epsilon=8/255$. During training, we use PGD~\cite{madry2017towards} to generate adversarial examples with step $K=10$ and step size $\eta=0.007$. We use a cohort of two networks in MAT. We set $
\alpha=0.6$ for CIFAR-10 and $\alpha=0.45$ for CIFAR-100.
\subsubsection{Comparisons with state-of-the-art}
\label{sec:sota}
We compare the performance of MAT trained models with state-of-the-art defenses including  MART~\cite{wang2019improving} and TRADES~\cite{zhang2019theoretically}, as well as vanilla AT~\cite{madry2017towards}. We evaluate the model robustness against FGSM~\cite{goodfellow2014explaining}, PGD~\cite{madry2017towards}, C\&W~\cite{carlini2017towards} and Fog, Elastic and Gabor attacks~\cite{kang2019testing}. For PGD, we use step $K=100$ and step size $\eta=0.003$. For C\&W attack~\cite{carlini2017towards}, the maximum number of iterations is set to 1,000 with a learning rate of 0.01. The evaluation results are summarized in Table~\ref{table:sota}. The two models trained in MAT outperform baseline models under all attacks considered, especially on the CIFAR-10 dataset. MAT improves model robustness without significant drops in clean accuracy. In fact, MAT models have higher accuracy in both clean and adversarial data on CIFAR-100 dataset compared to TRADES and MART.

Interestingly, the two models trained in MAT demonstrate different robustness characteristics: one model has slightly higher robust accuracy but lower clean accuracy, and the other has slightly higher clean accuracy and lower robust accuracy. In addition, transfer attacks from one model do not perform well on the other (Table~\ref{table:blackbox}). These results demonstrate that although the two models are trained collaboratively, they indeed learn different robust features and do not converge to one model.

\begin{table}
\centering
\caption{Classification accuracy (\%) of different defense methods. MAT-$h_1$ and MAT-$h_2$ are the two models trained in MAT. The best performance of each network architecture and dataset is in \textbf{bold} and the second best is \underline{underlined}.}
\label{table:sota}
\small
\scalebox{0.9}{
\begin{tabular}{c|ccccccccc}
\cline{1-10}
\multicolumn{2}{c}{Dataset}                  & Method   & Clean   & FGSM & PGD    & C\&W      & Fog     & Gabor   & Elastic  \\
\cline{1-10}
\multirow{10}{*}{\rotatebox{90}{ResNet-18}} & \multirow{5}{*}{CIFAR-10}  & AT~\cite{madry2017towards}       & \textbf{84.21} & 63.95 & 49.52 & 49.47 & 40.59 & 69.48 & 51.30  \\
                           & & TRADES~\cite{zhang2019theoretically}   & 81.48 & 62.60 & 52.26 & 49.92 & 38.48 & 68.08 & 51.61  \\
                          & & MART~\cite{wang2019improving}     & \underline{83.07} & 65.81 & 53.47 & 50.03 & 41.80 & 70.01 & 51.74  \\
                          & & MAT-$h_1$ & 81.22 & \textbf{66.53} & \textbf{57.17}  & \textbf{51.54} & \textbf{46.27} & \textbf{71.36} & {56.73}  \\
                          & & MAT-$h_2$ & 81.91 & \underline{66.45} & \underline{56.30} & \underline{51.12} & \underline{46.01} & \underline{71.19} & \textbf{57.11}  \\
\cline{2-10}
&\multirow{5}{*}{CIFAR-100} & AT~\cite{madry2017towards}        & \textbf{57.16} & 35.00 & 24.71 & 24.52 & 15.58 & 41.03 & 20.52  \\
                          & & TRADES~\cite{zhang2019theoretically}   & 54.04 & 35.33 & 28.26  & 24.63 & 14.63 & 40.89 & 23.03  \\
                          & & MART~\cite{wang2019improving}     & 54.44 & 38.40 & 31.85  & 27.81 & 20.46 & 43.66 & 26.07  \\
                          & & MAT-$h_1$ & \underline{56.06} & \underline{39.25} & \underline{31.99}  & \underline{28.32} & \underline{22.15} & \underline{44.90} & \textbf{28.37}  \\
                          & & MAT-$h_2$ & 55.98 & \textbf{39.69} & \textbf{32.54} & \textbf{28.42} & \textbf{22.47} & \textbf{45.32} & \underline{27.94} \\
\cline{1-10}
\multirow{10}{*}{\rotatebox{90}{WRN-34-10}}&\multirow{5}{*}{CIFAR-10}  & AT~\cite{madry2017towards}       & \textbf{86.50} & 59.06 & 50.72 & 51.88 & 42.88 & 71.21 & 49.94  \\
                         & & TRADES~\cite{zhang2019theoretically}   & 84.92 & 60.87 & 55.58 & 54.36 & 45.73 & 72.39 & 52.52  \\
                         & & MART~\cite{wang2019improving}    & 83.62 & 61.61 & 56.49 & 53.28 & 43.34 & 72.06 & 53.03  \\
                         & & MAT-$h_1$ & \underline{85.00} & \underline{64.20} & \textbf{59.02} & \underline{55.41} & \underline{49.24} & \underline{73.81} & \underline{54.65}  \\
                         & & MAT-$h_2$ & 84.96 & \textbf{64.23} & \underline{58.86} & \textbf{55.65} & \textbf{49.56} & \textbf{74.12} & \textbf{55.17}  \\
\cline{2-10}
&\multirow{5}{*}{CIFAR-100} & AT~\cite{madry2017towards}         & 60.76 & 36.38 & 25.74 & 26.49 & 17.05 & 43.44 & 22.21  \\
                         & & TRADES~\cite{zhang2019theoretically}   & 57.83 & 38.05 & 30.49 & 27.66 & 19.01 & 43.91 & 23.86  \\
                         & & MART~\cite{wang2019improving}     & 58.48 & \underline{41.40} & 33.07 &      29.85   & 22.07 & 46.55 & 24.89  \\
                         & & MAT-$h_1$ & \textbf{62.28} & \textbf{42.57} & \underline{33.63} & \underline{31.35} & \underline{24.97} & \textbf{48.71} & \underline{27.19}  \\
                         & & MAT-$h_2$ & \underline{62.20} & 41.36 & \textbf{33.78} & \textbf{31.38} & \textbf{25.81} & \underline{48.58} & \textbf{27.89} \\
\cline{1-10}
\end{tabular}}
\end{table}

\subsubsection{Comparisons with KD defenses}
\label{sec:kd}
We compare the performance of MAT trained models with other KD defenses including ARD~\cite{goldblum2019adversarially} and ACT~\cite{arani2020adversarial}. We evaluate the models using $K$ step PGD attacks (PGD-$K$) with step size $\eta=0.003$ and $K=20$, 100 and 1000 following~\cite{arani2020adversarial}. For ARD, we use the same network architecture for both the teacher model and student model, and report the performance of the student models. The results are summarized in Table~\ref{table:kd}. MAT models achieve significantly higher robustness compared to ARD and ACT. ARD does not perform very well compared to ACT and MAT, especially on the CIFAR-100 dataset.


\begin{table}
\centering
\small
\caption{Classification accuracy (\%) of different knowledge distillation defense methods. MAT-$h_1$ and MAT-$h_2$ are the two models trained in MAT. The best performance of each network architecture and dataset is in \textbf{bold} and the second best is \underline{underlined}.}
\label{table:kd}
\scalebox{0.83}{
\begin{tabular}{c|c|cccc|cccc}
\cline{1-10}
\multirow{2}{*}{Dataset}                   & \multirow{2}{*}{Method} & \multicolumn{4}{c|}{ResNet-18}    & \multicolumn{4}{c}{WRN-34-10} \\ \cline{3-10}

  & & Clean & PGD-20 & PGD-100 & PGD-1000 & Clean & PGD-20 & PGD-100 & PGD-1000  \\
\cline{1-10}
\multirow{5}{*}{CIFAR-10}  & AT~\cite{madry2017towards}       & \underline{84.21} & 51.66  & 49.50    & 49.40   & \underline{86.50} & 53.04  & 50.72    & 50.65   \\
                          & ARD~\cite{goldblum2019adversarially}      & 82.84 & 51.41  & 49.57   & 49.51   &  85.18  & 53.79    & 51.71 & 51.61 \\
                          & ACT~\cite{arani2020adversarial}      & \textbf{84.33} & 55.83  & 53.73   & 53.62    & \textbf{87.10} & 54.77 &  50.65   &  50.51\\
                          & MAT-$h_1$ & 81.22 & \textbf{58.67}  & \textbf{57.17}   & \textbf{57.13}   & 85.00 & \underline{60.52}  & \textbf{59.02}   & \textbf{58.93} \\
                          & MAT-$h_2$ & 81.76 & \underline{58.01}  & \underline{56.30}   & \underline{56.28}   & 84.96 & \textbf{60.68}  & \underline{58.86}    & \underline{58.84}  \\
\cline{1-10}
\multirow{5}{*}{CIFAR-100} & AT~\cite{madry2017towards}       & \underline{57.16} & 26.13  & 24.71   & 24.66   & 60.76 & 27.09  & 25.74   & 25.72  \\
                          & ARD~\cite{goldblum2019adversarially}      & 47.72 & 25.29  & 24.57   & 24.54   & 45.89 & 25.16 &  23.97  & 23.93 \\
                          & ACT~\cite{arani2020adversarial}      & \textbf{60.72} & 28.74  & 27.32   & 27.26   & 61.84 & 28.78  & 26.66    &  26.56 \\
                          & MAT-$h_1$ & 56.06 & \underline{32.68}  & \underline{31.99}   & \underline{31.98}  & \textbf{62.28} & \textbf{34.74}  & \underline{33.64}    &  \underline{33.63}  \\
                          & MAT-$h_2$ & 55.98 & \textbf{33.29}  & \textbf{32.54 }  & \textbf{32.52}  & \underline{62.20} & \underline{34.67}  & \textbf{33.78}   & \textbf{33.73}   \\
\cline{1-10}
\end{tabular}}
\end{table}

\subsubsection{Gradient Obfuscation}
We further verify that the robustness of MAT does not come from gradient obfuscation~\cite{athalye2018obfuscated}. This claim is supported by two observations: (1) One-step attacks, \eg, FGSM, perform worse than iterative attacks, \eg, PGD (see Table~\ref{table:sota}); (2) white-box attacks have higher success rates than black-box attacks (see Table~\ref{table:blackbox}). 
\begin{table}
\centering
\small
\caption{Classification accuracy (\%) of MAT models under PGD black-box attacks from different source models. We use ResNet-18 networks and PGD attacks with step $K=100$ and step size $\eta=0.003$.}
\label{table:blackbox}
\scalebox{0.9}{
\begin{tabular}{ccccccc}
\toprule
Dataset                   & Model    & AT    & TRADES & MART  & MAT-$h_1$ & MAT-$h_2$  \\
\midrule
\multirow{2}{*}{CIFAR-10}  & MAT-$h_1$ & 62.74 & 62.50   & 63.10  & 57.16    & 63.49     \\
                          & MAT-$h_2$ & 62.56 & 62.11  & 63.03 & 63.92    & 56.35     \\
\midrule
\multirow{2}{*}{CIFAR-100} & MAT-$h_1$ & 39.49 & 40.62  & 38.11 & 31.99    & 37.61     \\
                          & MAT-$h_2$ & 39.96 & 41.41  & 38.51 & 38.36    & 32.41    \\
\bottomrule
\end{tabular}}
\end{table}

\subsubsection{Different Scenarios of MAT}
\begin{figure}
    \centering
    \subfigure[]
    {
        \includegraphics[scale=.4]{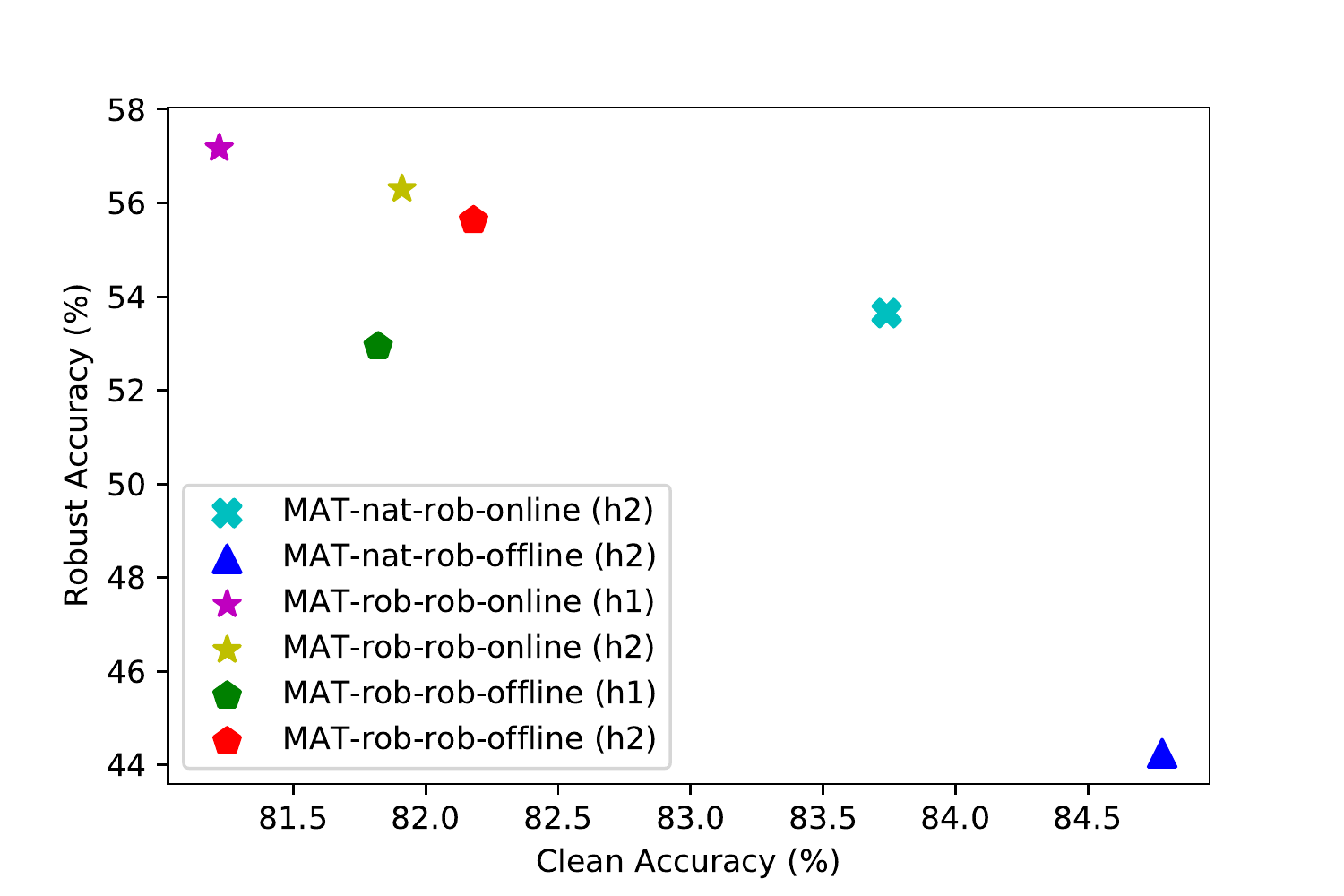}
    }
    \subfigure[]
    {
        \includegraphics[scale=.4]{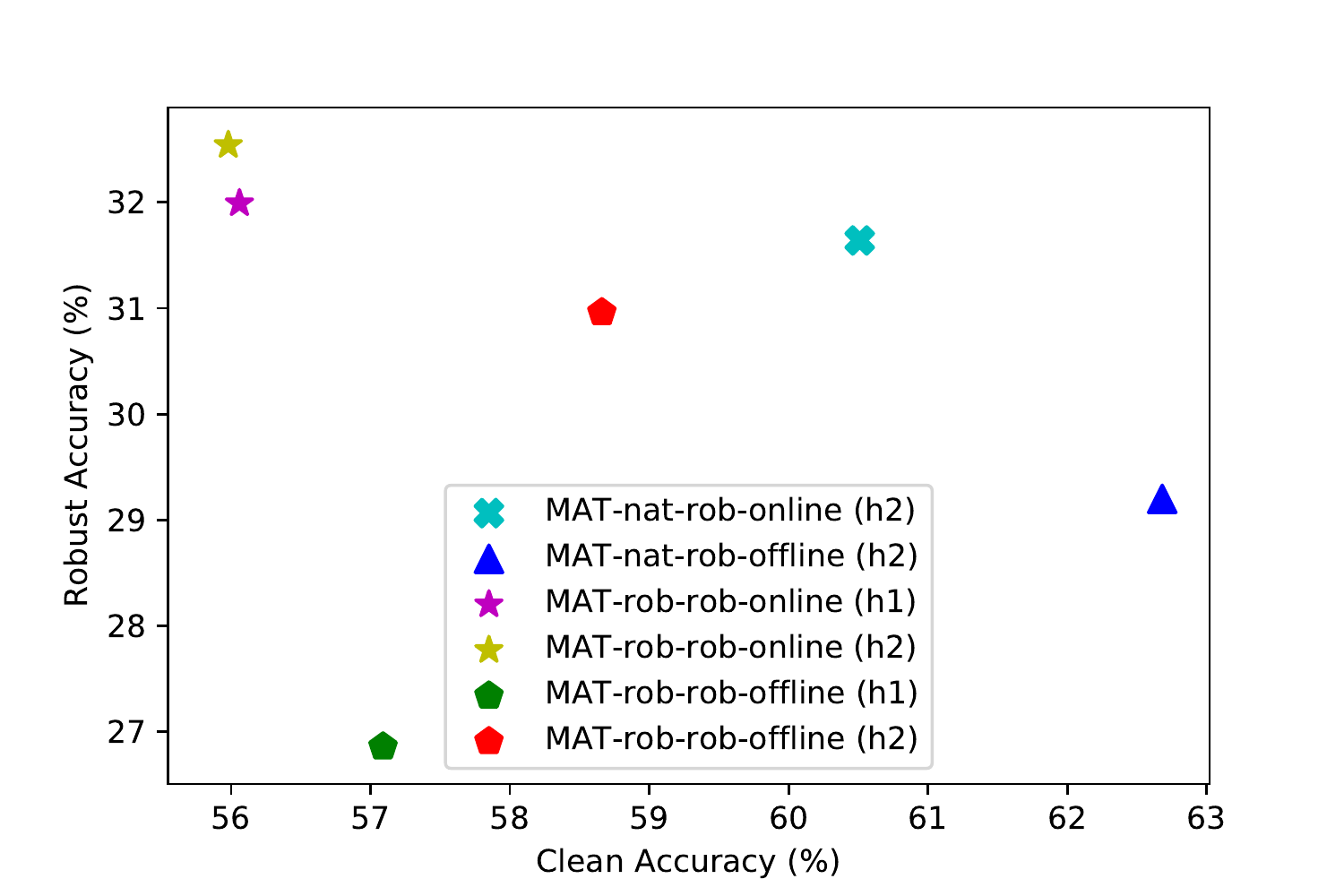}
    }
    \caption
    { Classification accuracy of ResNet-18 models trained in different MAT scenarios. Robust accuracy is the classification accuracy under PGD-100 attack. 
        (a) CIFAR-10 results;
        (b) CIFAR-100 results.
    }
    \label{fig:peer}
\end{figure}

In this section, we investigate how the interactions among the models affect the robustness behavior in the MAT algorithm, where models exchange their knowledge via KD. We consider four different KD scenarios in MAT: 1) MAT-rob-rob-online: $h_1$ and $h_2$ are both robust models and are trained simultaneously using Eq.(\ref{eq:mat-loss}), which is the scenario considered in this paper; 2) MAT-rob-rob-offline: $h_1$ and $h_2$ are both robust models, $h_1$ is a teacher model trained by Eq.(\ref{eq:at1}) and $h_2$ is a student model learning from $h_1$ using Eq.(\ref{eq:mat-loss}) with $h_1$ fixed; 3) MAT-nat-rob-online: $h_1$ is a natural model and $h_2$ is a robust model, and they are trained simultaneously using Eq.(\ref{eq:mat-loss}) with $\xadv{\thetaa}$ replaced by $x$; 4) MAT-nat-rob-offline: $h_1$ is a natural model and $h_2$ is a robust model, $h_1$ is a teacher model trained by $\loss{C}$, and $h_2$ is a student model learning from $h_1$ using Eq.(\ref{eq:mat-loss}) with $h_1$ fixed and $\xadv{\thetaa}$ replaced by $x$. Fig.~\ref{fig:peer} shows both robust and clean accuracies corresponding to the four scenarios. We observe that 1) robust models trained with natural models have high clean accuracy; 2) robust models trained with robust models have high robust accuracy; 3) MAT-rob-rob-online achieves higher robust accuracy than MAT-rob-rob-offline, which confirms our intuition: by learning from peers, a network becomes more robust, which in turn improves the robustness of its peers, while in MAT-rob-rob-offline the teacher model is fixed which limits the performance of the student. We use MAT-rob-rob-online in this paper as it achieves the highest robust accuracy. 

\subsubsection{Effect of $\alpha$}
\label{sec:hp}
\begin{figure}
    \centering
    \subfigure[]
    {
        \includegraphics[scale=.4]{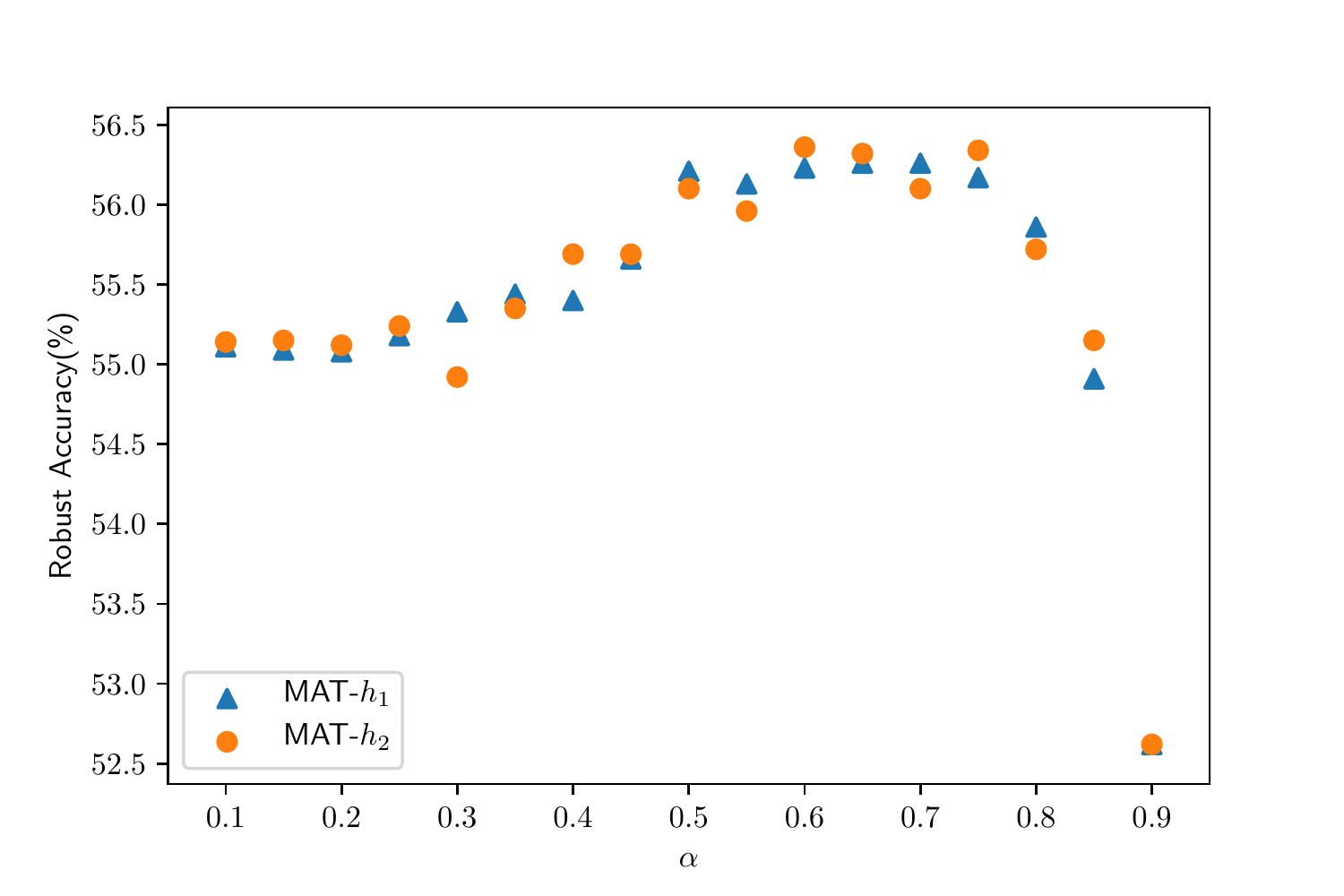}
    }
    \subfigure[]
    {
        \includegraphics[scale=.4]{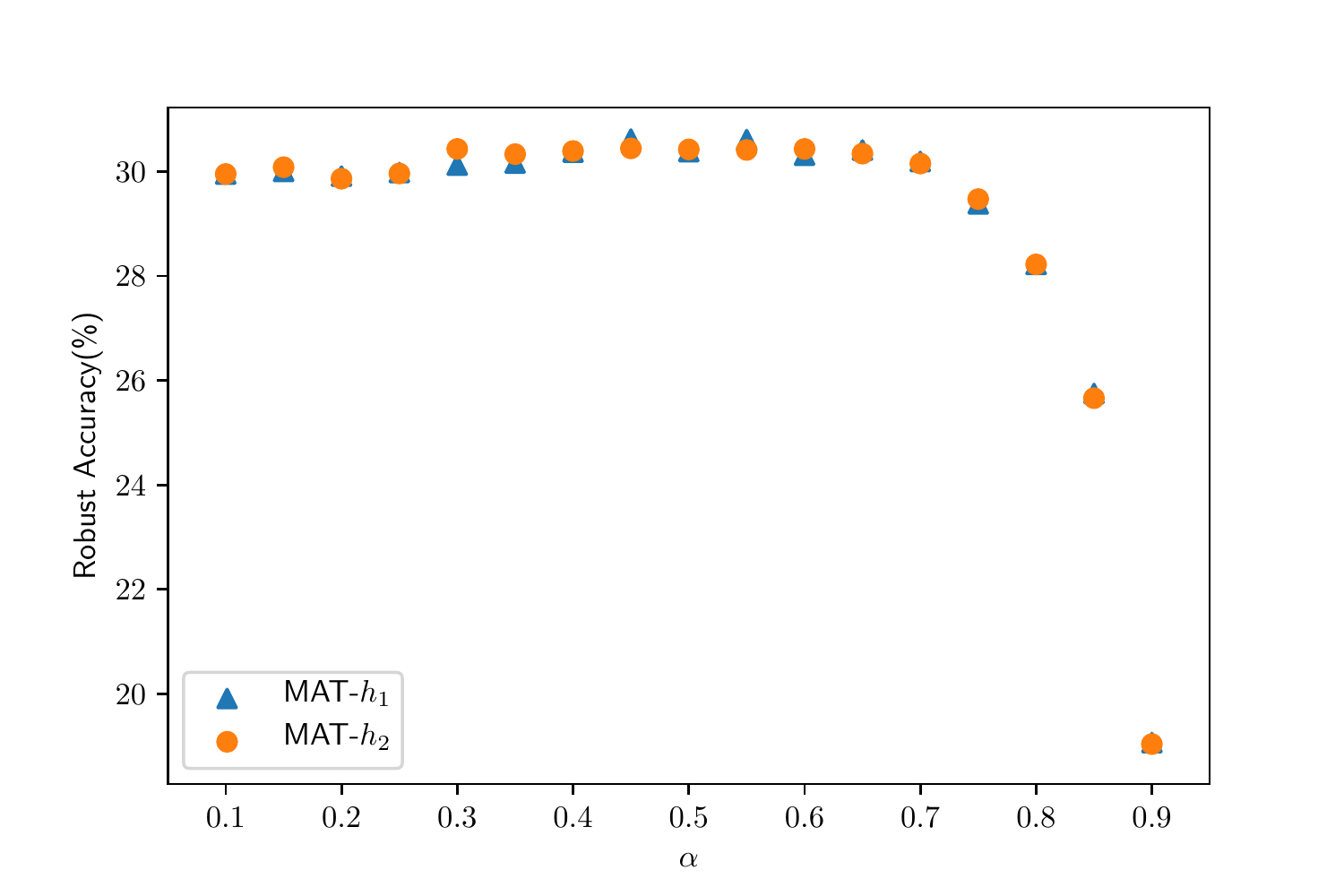}
    }
    \caption
    { Performance of ResNet-18 models trained with different $\alpha$ values. Robust accuracy is the classification accuracy under PGD-20 attack on the validation set. 
        (a) CIFAR-10 results;
        (b) CIFAR-100 results.
    }
    \label{fig:hp}
\end{figure}

The hyper-parameter $\alpha$ in Eq.~\ref{eq:mat-loss} controls the trade-off between learning from ground-truth labels and peer network. In order to search for the optimal $\alpha$, we randomly select 20\% of CIFAR-10 and CIFAR-100 training sets as the validation sets, and train ResNet-18 models using different values of $\alpha$ on the other 80\% of training data. Fig.~\ref{fig:hp} shows the effect of $\alpha$ on CIFAR-10 and CIFAR-100 datasets. In general, the robustness of the models increases with $\alpha$ as the models start to learn more from each other, and drops rapidly when $\alpha$ becomes too high as the models learns too little from ground-truth labels. In addition, MAT is less sensitive to the value of $\alpha$ on CIFAR-100 compared to CIFAR-10. We choose $\alpha=0.6$ for CIFAR-10 and $\alpha=0.45$ for CIFAR-100 as they achieve the highest robust accuracy on the validation sets. After determining the values of $\alpha$, we use the full training sets for training. 
\subsection{Defense against Multiple Perturbations}
\subsubsection{Settings} 
\label{sec:mp-settings}
In this section, we evaluate the effectiveness of MAT-MP. We consider three perturbation types: $l_\infty$, $l_2$ and $l_1$ attacks with $\epsilon=(0.03, 0.5, 12)$. We use ResNet-18~\cite{he2016deep} networks. During training, we train three specialists using PGD $l_\infty$, $l_2$, and $l_1$ attacks with step size $\eta=(0.003, 0.05,0.05)$, number of iterations $K=(40, 50, 50)$ respectively, and one generalist that aims to defend against all perturbation types by learning from the specialists. 

During evaluation, a broad suite of both gradient and non-gradient based attacks are used for each perturbation type. $l_\infty$ attacks include FGSM, PGD, MIM~\cite{dong2018boosting}; $l_2$ attacks include PGD, the Gaussian noise attack~\cite{rauber2017foolbox}, the boundary attack~\cite{brendel2017decision}, DeepFool~\cite{moosavi2016deepfool}, the pointwise attack~\cite{schott2018towards}, DDN attack~\cite{rony2019decoupling}, and C\&W attack~\cite{carlini2017towards}; $l_1$ attacks include PGD, the salt \& pepper attack~\cite{rauber2017foolbox}, and the pointwise attack~\cite{schott2018towards}. Each attack is run with 10 random starts. The evaluation details can be found in~\cite{maini2020adversarial}.

\subsubsection{Results}
 We compare the robustness of our generalist models with models trained with PGD $l_p$ attacks only (AT-$l_p$), model trained with the worst PGD attack (AT-MAX)~\cite{tramer2019adversarial, maini2020adversarial}, model trained with all PGD attacks (AT-AVG)~\cite{tramer2019adversarial, maini2020adversarial}, and model trained with MSD attacks (AT-MSD)~\cite{maini2020adversarial}. Baseline models are provided by the authors of~\cite{maini2020adversarial}. The results are summarized in Table~\ref{table:mp-all}. We report the average classification accuracy under three perturbation types $\mathcal{R}_{avg}$, as well as the worst-case accuracy $\mathcal{R}_{worst}$ where we pick the strongest attack from all attacks considered for each example. We can make several observations from Table~\ref{table:mp-all}: 

1) \textit{MAT produces the strongest model against multiple perturbations.} MAT-AVG model achieves the best average accuracy against multiple perturbations, which reaches 58.7\% average accuracy against $l_\infty$, $l_2$, and $l_1$ attacks (individually 47.3\%, 68.9\%, 60.0\%); MAT-MAX model achieves the best performance against the union of attacks, which reaches 48.0\% accuracy against the union of $l_\infty$, $l_2$, and $l_1$ attacks (individually 51.1\%, 67.0\%, 54.0\%). 

2) \textit{MAT improves model robustness as well as clean accuracy.} Compared to their AT counterparts, MAT trained models have higher clean accuracy and adversarial accuracy under all perturbation types. This demonstrates the efficacy of learning from the specialists models. Improvement is most significant when using "MAX" training strategy: MAT-MAX achieves 3.5\% higher clean accuracy, 8.7\% higher $\mathcal{R}_{avg}$ and 13.1\% higher $\mathcal{R}_{worst}$ than AT-MAX. This may be because simple generalization of adversarial training such as "MAX" converges to sub-optimal solutions that are unable to balance the right trade-off between multiple attacks~\cite{tramer2019adversarial, maini2020adversarial}. By learning from the soft labels provided by specialists models, the generalist is aware of the optimal decision boundary of each perturbation type and find a better trade-off between multiple perturbations. 

3) Among MAT models, MAT-MAX has the highest $\mathcal{R}_{worst}$ and MAT-AVG has the highest $\mathcal{R}_{avg}$, which is unsurprising since MAT-MAX model is trained on the worst PGD $l_p$ attack and MAT-AVG model is trained on all PGD $l_p$ attacks. MAT-MSD achieves higher $\mathcal{R}_{worst}$ than MAT-AVG, which indicates that it has better robustness against the union of multiple perturbations, but it has slightly worse $\mathcal{R}_{avg}$ and $\mathcal{R}_{worst}$ than MAT-MAX. 

In summary, our results suggest that by learning from specialists models, the generalist is able to inherit their robustness against different perturbations and find a decision boundary that is more robust and more generalizable against multiple perturbations. 

\begin{table}
\centering
\caption{Summary of classification accuracy results on CIFAR-10.  $\mathcal{R}_{avg}$ is the average accuracy of different perturbation types. $\mathcal{R}_{worst}$ is the worst-case accuracy. We show the accuracy gain of MAT models over their AT counterparts in the brackets. The best performance in each column is in \textbf{bold} and the second best is \underline{underlined}.}
\label{table:mp-all}
\small
\scalebox{0.9}{
\begin{tabular}{ccccccc}
\hline
Model   & Clean  & $l_\infty$ attacks  & $l_2$ attacks & $l_1$ attacks  & $\mathcal{R}_{avg}$    & $\mathcal{R}_{worst}$     \\\hline
AT-$l_\infty$~\cite{maini2020adversarial}    & 83.3\% & \underline{50.7\%} & 57.3\% & 16.0\% & 41.3\% & 15.6\%  \\
AT-$l_2$~\cite{maini2020adversarial}    & \textbf{90.2\%} & 28.3\% & 61.6\% & 46.6\% & 45.5\% & 27.5\%  \\
AT-$l_1$~\cite{maini2020adversarial}    & 73.3\% & 0.2\%  & 0.0\%  & 7.9\%  & 2.7\%  & 0.0\%   \\\hline
AT-MAX~\cite{maini2020adversarial}     & 81.0\% & 44.9\% & 61.7\% & 39.4\% & 48.7\% & 34.9\%  \\
MAT-MAX \textbf{(Ours)} & 84.5\% & \textbf{51.1\%} & 67.0\% & 54.0\% & \underline{57.4\%} & \textbf{48.0\%}  \\
                         & {(+3.5\%)}  & {(+6.2\%) }                  & {(+5.3\%)}  & {(+14.6\%)} & {(+8.7\%)}  & {(+13.1\%)}  \\\hline
AT-AVG~\cite{maini2020adversarial}     & 84.6\% & 42.5\% & 65.0\% & 54.0\% & 53.8\% & 40.6\%  \\
MAT-AVG \textbf{(Ours)}& \underline{86.0\%} & 47.3\% & \textbf{68.9\%} & \textbf{60.0\%} & \textbf{58.7\%} & 45.9\%  \\                         & {(+1.4\%)}  & {(+4.8\%) }                  & {(+3.9\%)}  & {(+6.0\%)} & {(+4.9\%)}  & {(+5.3\%)}  \\\hline
AT-MSD~\cite{maini2020adversarial}     & 81.1\% & 48.0\% & 64.3\% & 53.0\% & 55.1\% & 47.0\%  \\
MAT-MSD \textbf{(Ours)} & 84.2\% & 48.8\% & \underline{67.6\%} & \underline{55.1\%} & 57.2\% & \underline{47.7\%} \\                         & {(+3.1\%)}  & {(+0.8\%) }                  & {(+3.3\%)}  & {(+2.1\%)} & {(+2.1\%)}  & {(+0.7\%)}  \\\hline
\end{tabular}}
\end{table}

\section{Conclusion}
In this paper, we proposed a novel knowledge distillation based AT algorithm named MAT that allows models to learn from each other during adversarial training. MAT is a general framework and can be utilized to defend against multiple perturbations. We demonstrated that the proposed method can effectively improve model robustness and outperform state-of-the-art methods in CIFAR-10 and CIFAR-100 datasets. Our results demonstrate that collaborative learning is an effective strategy for training robust models.
\section*{Acknowledgement} 
This work was supported by the DARPA GARD Program HR001119S0026-GARD-FP-052.
\bibliography{main}
\end{document}

%% file: math_commands.tex
\usepackage{amsmath,amsfonts,bm}

\def\eqref#1{equation~\ref{#1}}

\def\1{\bm{1}}

\DeclareMathAlphabet{\mathsfit}{\encodingdefault}{\sfdefault}{m}{sl}
\SetMathAlphabet{\mathsfit}{bold}{\encodingdefault}{\sfdefault}{bx}{n}

\def\gX{{\mathcal{X}}}
\def\gY{{\mathcal{Y}}}

\def\sR{{\mathbb{R}}}

\DeclareMathOperator*{\argmax}{arg\,max}

%% file: main.bbl
\begin{thebibliography}{47}
\providecommand{\natexlab}[1]{#1}
\providecommand{\url}[1]{\texttt{#1}}
\expandafter\ifx\csname urlstyle\endcsname\relax
  \providecommand{\doi}[1]{doi: #1}\else
  \providecommand{\doi}{doi: \begingroup \urlstyle{rm}\Url}\fi

\bibitem[Arani et~al.(2020)Arani, Sarfraz, and Zonooz]{arani2020adversarial}
Elahe Arani, Fahad Sarfraz, and Bahram Zonooz.
\newblock Adversarial concurrent training: Optimizing robustness and accuracy
  trade-off of deep neural networks.
\newblock \emph{arXiv preprint arXiv:2008.07015}, 2020.

\bibitem[Athalye et~al.(2018)Athalye, Carlini, and
  Wagner]{athalye2018obfuscated}
Anish Athalye, Nicholas Carlini, and David Wagner.
\newblock Obfuscated gradients give a false sense of security: Circumventing
  defenses to adversarial examples.
\newblock In \emph{International Conference on Machine Learning}, pages
  274--283. PMLR, 2018.

\bibitem[Brendel et~al.(2017)Brendel, Rauber, and Bethge]{brendel2017decision}
Wieland Brendel, Jonas Rauber, and Matthias Bethge.
\newblock Decision-based adversarial attacks: Reliable attacks against
  black-box machine learning models.
\newblock \emph{arXiv preprint arXiv:1712.04248}, 2017.

\bibitem[Carlini and Wagner(2016)]{carlini2016defensive}
Nicholas Carlini and David Wagner.
\newblock Defensive distillation is not robust to adversarial examples.
\newblock \emph{arXiv preprint arXiv:1607.04311}, 2016.

\bibitem[Carlini and Wagner(2017)]{carlini2017towards}
Nicholas Carlini and David Wagner.
\newblock Towards evaluating the robustness of neural networks.
\newblock In \emph{2017 IEEE Symposium on Security and Privacy (SP)}, pages
  39--57. IEEE, 2017.

\bibitem[Chen et~al.(2017)Chen, Sharma, Zhang, Yi, and Hsieh]{chen2017ead}
Pin-Yu Chen, Yash Sharma, Huan Zhang, Jinfeng Yi, and Cho-Jui Hsieh.
\newblock {EAD}: {E}lastic-net attacks to deep neural networks via adversarial
  examples.
\newblock \emph{arXiv preprint arXiv:1709.04114}, 2017.

\bibitem[Chen et~al.(2021)Chen, Zhang, Liu, Chang, and Wang]{chen2021robust}
Tianlong Chen, Zhenyu Zhang, Sijia Liu, Shiyu Chang, and Zhangyang Wang.
\newblock Robust overfitting may be mitigated by properly learned smoothening.
\newblock In \emph{International Conference on Learning Representations}, 2021.
\newblock URL \url{https://openreview.net/forum?id=qZzy5urZw9}.

\bibitem[Cohen et~al.(2019)Cohen, Rosenfeld, and Kolter]{cohen2019certified}
Jeremy Cohen, Elan Rosenfeld, and Zico Kolter.
\newblock Certified adversarial robustness via randomized smoothing.
\newblock In \emph{International Conference on Machine Learning}, pages
  1310--1320. PMLR, 2019.

\bibitem[Dong et~al.(2018)Dong, Liao, Pang, Su, Zhu, Hu, and
  Li]{dong2018boosting}
Yinpeng Dong, Fangzhou Liao, Tianyu Pang, Hang Su, Jun Zhu, Xiaolin Hu, and
  Jianguo Li.
\newblock Boosting adversarial attacks with momentum.
\newblock In \emph{Proceedings of the IEEE {C}onference on {C}omputer {V}ision
  and {P}attern {R}ecognition}, pages 9185--9193, 2018.

\bibitem[Furlanello et~al.(2018)Furlanello, Lipton, Tschannen, Itti, and
  Anandkumar]{furlanello2018born}
Tommaso Furlanello, Zachary Lipton, Michael Tschannen, Laurent Itti, and Anima
  Anandkumar.
\newblock Born again neural networks.
\newblock In \emph{International Conference on Machine Learning}, pages
  1607--1616. PMLR, 2018.

\bibitem[Goldblum et~al.(2019)Goldblum, Fowl, Feizi, and
  Goldstein]{goldblum2019adversarially}
Micah Goldblum, Liam Fowl, Soheil Feizi, and Tom Goldstein.
\newblock Adversarially robust distillation.
\newblock \emph{arXiv preprint arXiv:1905.09747}, 2019.

\bibitem[Goodfellow et~al.(2014)Goodfellow, Shlens, and
  Szegedy]{goodfellow2014explaining}
Ian~J Goodfellow, Jonathon Shlens, and Christian Szegedy.
\newblock Explaining and harnessing adversarial examples.
\newblock \emph{arXiv preprint arXiv:1412.6572}, 2014.

\bibitem[Guo et~al.(2020)Guo, Wang, Wu, Yu, Liang, Hu, and Luo]{guo2020online}
Qiushan Guo, Xinjiang Wang, Yichao Wu, Zhipeng Yu, Ding Liang, Xiaolin Hu, and
  Ping Luo.
\newblock Online knowledge distillation via collaborative learning.
\newblock In \emph{Proceedings of the IEEE {C}onference on {C}omputer {V}ision
  and {P}attern {R}ecognition}, pages 11020--11029, 2020.

\bibitem[He et~al.(2016)He, Zhang, Ren, and Sun]{he2016deep}
Kaiming He, Xiangyu Zhang, Shaoqing Ren, and Jian Sun.
\newblock Deep residual learning for image recognition.
\newblock In \emph{Proceedings of the IEEE {C}onference on {C}omputer {V}ision
  and {P}attern {R}ecognition}, pages 770--778, 2016.

\bibitem[{Hinton} et~al.(2012){Hinton}, {Deng}, {Yu}, {Dahl}, {Mohamed},
  {Jaitly}, {Senior}, {Vanhoucke}, {Nguyen}, {Sainath}, and
  {Kingsbury}]{6296526}
G.~{Hinton}, L.~{Deng}, D.~{Yu}, G.~E. {Dahl}, A.~{Mohamed}, N.~{Jaitly},
  A.~{Senior}, V.~{Vanhoucke}, P.~{Nguyen}, T.~N. {Sainath}, and
  B.~{Kingsbury}.
\newblock Deep neural networks for acoustic modeling in speech recognition: The
  shared views of four research groups.
\newblock \emph{IEEE Signal Processing Magazine}, 29\penalty0 (6):\penalty0
  82--97, 2012.
\newblock \doi{10.1109/MSP.2012.2205597}.

\bibitem[Hinton et~al.(2015)Hinton, Vinyals, and Dean]{hinton2015distilling}
Geoffrey Hinton, Oriol Vinyals, and Jeff Dean.
\newblock Distilling the knowledge in a neural network.
\newblock \emph{arXiv preprint arXiv:1503.02531}, 2015.

\bibitem[Kang et~al.(2019)Kang, Sun, Hendrycks, Brown, and
  Steinhardt]{kang2019testing}
Daniel Kang, Yi~Sun, Dan Hendrycks, Tom Brown, and Jacob Steinhardt.
\newblock Testing robustness against unforeseen adversaries.
\newblock \emph{arXiv preprint arXiv:1908.08016}, 2019.

\bibitem[Kannan et~al.(2018)Kannan, Kurakin, and
  Goodfellow]{Kannan2018AdversarialLP}
Harini Kannan, A.~Kurakin, and Ian~J. Goodfellow.
\newblock Adversarial logit pairing.
\newblock \emph{ArXiv}, abs/1803.06373, 2018.

\bibitem[Krizhevsky(2012)]{krizhevsky2009learning}
Alex Krizhevsky.
\newblock Learning multiple layers of features from tiny images.
\newblock \emph{University of Toronto}, 05 2012.

\bibitem[Krizhevsky et~al.(2012)Krizhevsky, Sutskever, and
  Hinton]{krizhevsky2012imagenet}
Alex Krizhevsky, Ilya Sutskever, and Geoffrey~E Hinton.
\newblock Imagenet classification with deep convolutional neural networks.
\newblock In \emph{Advances in Neural Information Processing Systems}, pages
  1097--1105, 2012.

\bibitem[Laidlaw and Feizi(2019)]{laidlaw2019functional}
Cassidy Laidlaw and Soheil Feizi.
\newblock Functional adversarial attacks.
\newblock \emph{arXiv preprint arXiv:1906.00001}, 2019.

\bibitem[Laidlaw et~al.(2020)Laidlaw, Singla, and Feizi]{laidlaw2020perceptual}
Cassidy Laidlaw, Sahil Singla, and Soheil Feizi.
\newblock Perceptual adversarial robustness: Defense against unseen threat
  models.
\newblock \emph{arXiv preprint arXiv:2006.12655}, 2020.

\bibitem[Levine and Feizi(2020)]{levine2020randomized}
Alexander Levine and Soheil Feizi.
\newblock (de) randomized smoothing for certifiable defense against patch
  attacks.
\newblock \emph{arXiv preprint arXiv:2002.10733}, 2020.

\bibitem[Levine and Feizi(2021)]{levine2021improved}
Alexander Levine and Soheil Feizi.
\newblock Improved, deterministic smoothing for l1 certified robustness.
\newblock \emph{arXiv preprint arXiv:2103.10834}, 2021.

\bibitem[Levine et~al.(2019)Levine, Singla, and Feizi]{levine2019certifiably}
Alexander Levine, Sahil Singla, and Soheil Feizi.
\newblock Certifiably robust interpretation in deep learning.
\newblock \emph{arXiv preprint arXiv:1905.12105}, 2019.

\bibitem[Levine et~al.(2018)Levine, Pastor, Krizhevsky, Ibarz, and
  Quillen]{levine2018learning}
Sergey Levine, Peter Pastor, Alex Krizhevsky, Julian Ibarz, and Deirdre
  Quillen.
\newblock Learning hand-eye coordination for robotic grasping with deep
  learning and large-scale data collection.
\newblock \emph{The International Journal of Robotics Research}, 37\penalty0
  (4-5):\penalty0 421--436, 2018.

\bibitem[Lin et~al.(2020)Lin, Lau, Levine, Chellappa, and Feizi]{lin2020dual}
Wei-An Lin, Chun~Pong Lau, Alexander Levine, Rama Chellappa, and Soheil Feizi.
\newblock Dual {M}anifold {A}dversarial {R}obustness: {D}efense against {L}p
  and non-{L}p {A}dversarial {A}ttacks.
\newblock In \emph{Advances in Neural Information Processing Systems}, 2020.

\bibitem[Madry et~al.(2017)Madry, Makelov, Schmidt, Tsipras, and
  Vladu]{madry2017towards}
Aleksander Madry, Aleksandar Makelov, Ludwig Schmidt, Dimitris Tsipras, and
  Adrian Vladu.
\newblock Towards deep learning models resistant to adversarial attacks.
\newblock \emph{arXiv preprint arXiv:1706.06083}, 2017.

\bibitem[Maini et~al.(2020)Maini, Wong, and Kolter]{maini2020adversarial}
Pratyush Maini, Eric Wong, and Zico Kolter.
\newblock Adversarial robustness against the union of multiple perturbation
  models.
\newblock In \emph{International Conference on Machine Learning}, pages
  6640--6650. PMLR, 2020.

\bibitem[Moosavi-Dezfooli et~al.(2016)Moosavi-Dezfooli, Fawzi, and
  Frossard]{moosavi2016deepfool}
Seyed-Mohsen Moosavi-Dezfooli, Alhussein Fawzi, and Pascal Frossard.
\newblock Deepfool: a simple and accurate method to fool deep neural networks.
\newblock In \emph{Proceedings of the IEEE {C}onference on {C}omputer {V}ision
  and {P}attern {R}ecognition}, pages 2574--2582, 2016.

\bibitem[{Papernot} et~al.(2016){Papernot}, {McDaniel}, {Jha}, {Fredrikson},
  {Celik}, and {Swami}]{7467366}
N.~{Papernot}, P.~{McDaniel}, S.~{Jha}, M.~{Fredrikson}, Z.~B. {Celik}, and
  A.~{Swami}.
\newblock The limitations of deep learning in adversarial settings.
\newblock In \emph{2016 IEEE European Symposium on Security and Privacy (EuroS
  P)}, pages 372--387, 2016.
\newblock \doi{10.1109/EuroSP.2016.36}.

\bibitem[Papernot et~al.(2016)Papernot, McDaniel, Wu, Jha, and
  Swami]{papernot2016distillation}
Nicolas Papernot, Patrick McDaniel, Xi~Wu, Somesh Jha, and Ananthram Swami.
\newblock Distillation as a defense to adversarial perturbations against deep
  neural networks.
\newblock In \emph{2016 IEEE Symposium on Security and Privacy (SP)}, pages
  582--597. IEEE, 2016.

\bibitem[Raghunathan et~al.(2018)Raghunathan, Steinhardt, and
  Liang]{raghunathan2018certified}
Aditi Raghunathan, Jacob Steinhardt, and Percy Liang.
\newblock Certified defenses against adversarial examples.
\newblock In \emph{International Conference on Learning Representations}, 2018.
\newblock URL \url{https://openreview.net/forum?id=Bys4ob-Rb}.

\bibitem[Rauber et~al.(2017)Rauber, Brendel, and Bethge]{rauber2017foolbox}
Jonas Rauber, Wieland Brendel, and Matthias Bethge.
\newblock Foolbox: A python toolbox to benchmark the robustness of machine
  learning models.
\newblock \emph{arXiv preprint arXiv:1707.04131}, 2017.

\bibitem[Rony et~al.(2019)Rony, Hafemann, Oliveira, Ayed, Sabourin, and
  Granger]{rony2019decoupling}
J{\'e}r{\^o}me Rony, Luiz~G Hafemann, Luiz~S Oliveira, Ismail~Ben Ayed, Robert
  Sabourin, and Eric Granger.
\newblock Decoupling direction and norm for efficient gradient-based l2
  adversarial attacks and defenses.
\newblock In \emph{Proceedings of IEEE {C}onference on {C}omputer {V}ision and
  {P}attern {R}ecognition}, pages 4322--4330, 2019.

\bibitem[Schott et~al.(2019)Schott, Rauber, Bethge, and
  Brendel]{schott2018towards}
Lukas Schott, Jonas Rauber, Matthias Bethge, and Wieland Brendel.
\newblock Towards the first adversarially robust neural network model on
  {MNIST}.
\newblock In \emph{International Conference on Learning Representations}, 2019.
\newblock URL \url{https://openreview.net/forum?id=S1EHOsC9tX}.

\bibitem[Singla and Feizi(2021)]{singla2021skew}
Sahil Singla and Soheil Feizi.
\newblock Skew orthogonal convolutions.
\newblock \emph{arXiv preprint arXiv:2105.11417}, 2021.

\bibitem[Sinha et~al.(2017)Sinha, Namkoong, Volpi, and
  Duchi]{sinha2017certifying}
Aman Sinha, Hongseok Namkoong, Riccardo Volpi, and John Duchi.
\newblock Certifying some distributional robustness with principled adversarial
  training.
\newblock \emph{arXiv preprint arXiv:1710.10571}, 2017.

\bibitem[Song and Chai(2018)]{song2018collaborative}
Guocong Song and Wei Chai.
\newblock Collaborative learning for deep neural networks.
\newblock In \emph{Advances in Neural Information Processing Systems}, pages
  1832--1841, 2018.

\bibitem[Tram{\`e}r and Boneh(2019)]{tramer2019adversarial}
Florian Tram{\`e}r and Dan Boneh.
\newblock Adversarial training and robustness for multiple perturbations.
\newblock In \emph{Advances in Neural Information Processing Systems}, pages
  5866--5876, 2019.

\bibitem[Tram{\`e}r et~al.(2017)Tram{\`e}r, Papernot, Goodfellow, Boneh, and
  McDaniel]{tramer2017space}
Florian Tram{\`e}r, Nicolas Papernot, Ian Goodfellow, Dan Boneh, and Patrick
  McDaniel.
\newblock The space of transferable adversarial examples.
\newblock \emph{arXiv preprint arXiv:1704.03453}, 2017.

\bibitem[Tsipras et~al.(2019)Tsipras, Santurkar, Engstrom, Turner, and
  Madry]{tsipras2018robustness}
Dimitris Tsipras, Shibani Santurkar, Logan Engstrom, Alexander Turner, and
  Aleksander Madry.
\newblock Robustness may be at odds with accuracy.
\newblock In \emph{International Conference on Learning Representations}, 2019.
\newblock URL \url{https://openreview.net/forum?id=SyxAb30cY7}.

\bibitem[Wang and Yoon(2020)]{wang2020knowledge}
Lin Wang and Kuk-Jin Yoon.
\newblock Knowledge distillation and student-teacher learning for visual
  intelligence: A review and new outlooks.
\newblock \emph{arXiv preprint arXiv:2004.05937}, 2020.

\bibitem[Wang et~al.(2019)Wang, Zou, Yi, Bailey, Ma, and Gu]{wang2019improving}
Yisen Wang, Difan Zou, Jinfeng Yi, James Bailey, Xingjun Ma, and Quanquan Gu.
\newblock Improving adversarial robustness requires revisiting misclassified
  examples.
\newblock In \emph{International Conference on Learning Representations}, 2019.
\newblock URL \url{https://openreview.net/forum?id=rklOg6EFwS}.

\bibitem[Zagoruyko and Komodakis(2016)]{BMVC2016_87}
Sergey Zagoruyko and Nikos Komodakis.
\newblock Wide residual networks.
\newblock In Edwin R.~Hancock Richard C.~Wilson and William A.~P. Smith,
  editors, \emph{Proceedings of the British Machine Vision Conference (BMVC)},
  pages 87.1--87.12. BMVA Press, September 2016.
\newblock ISBN 1-901725-59-6.
\newblock \doi{10.5244/C.30.87}.
\newblock URL \url{https://dx.doi.org/10.5244/C.30.87}.

\bibitem[Zhang et~al.(2019)Zhang, Yu, Jiao, Xing, El~Ghaoui, and
  Jordan]{zhang2019theoretically}
Hongyang Zhang, Yaodong Yu, Jiantao Jiao, Eric Xing, Laurent El~Ghaoui, and
  Michael Jordan.
\newblock Theoretically principled trade-off between robustness and accuracy.
\newblock In \emph{International Conference on Machine Learning}, pages
  7472--7482, 2019.

\bibitem[Zhang et~al.(2018)Zhang, Xiang, Hospedales, and Lu]{zhang2018deep}
Ying Zhang, Tao Xiang, Timothy~M Hospedales, and Huchuan Lu.
\newblock Deep mutual learning.
\newblock In \emph{Proceedings of the IEEE {C}onference on {C}omputer {V}ision
  and {P}attern {R}ecognition}, pages 4320--4328, 2018.

\end{thebibliography}
